\pdfoutput=1
\documentclass[11pt]{article}
\usepackage{ACL2023}
\usepackage{times}
\usepackage{latexsym}
\usepackage[T1]{fontenc}
\usepackage[utf8]{inputenc}
\usepackage{microtype}
\usepackage{inconsolata}

\usepackage{stfloats}
\usepackage{booktabs}
\usepackage{hyperref}
\usepackage{graphicx}
\usepackage{multirow}
\usepackage{wrapfig}
\usepackage{placeins}
\usepackage{subcaption}
\usepackage[ruled]{algorithm2e}

\usepackage{amsmath,amsfonts,bm}

\def\eqref#1{equation~\ref{#1}}

\def\1{\bm{1}}

\DeclareMathAlphabet{\mathsfit}{\encodingdefault}{\sfdefault}{m}{sl}
\SetMathAlphabet{\mathsfit}{bold}{\encodingdefault}{\sfdefault}{bx}{n}

\DeclareMathOperator*{\argmin}{arg\,min}

\title{Towards Efficient Active Learning in NLP via Pretrained Representations}

\author{Artem Vysogorets$^\dagger$\thanks{$\dagger$ work performed while interning at Bloomberg}\\
  New York University \\
  Center for Data Science \\
  \texttt{amv458@nyu.edu} \\\And
  Achintya Gopal \\
  Bloomberg  \\
  731 Lexington Ave \\
  \texttt{agopal6@bloomberg.net} \\}

\begin{document}
\maketitle
\begin{abstract}
Fine-tuning Large Language Models (LLMs) is now a common approach for text classification in a wide range of applications. When labeled documents are scarce, active learning helps save annotation efforts but requires retraining of massive models on each acquisition iteration. We drastically expedite this process by using pretrained representations of LLMs within the active learning loop and, once the desired amount of labeled data is acquired, fine-tuning that or even a different pretrained LLM on this labeled data to achieve the best performance. As verified on common text classification benchmarks with pretrained BERT and RoBERTa as the backbone, our strategy yields similar performance to fine-tuning all the way through the active learning loop but is orders of magnitude less computationally expensive. The data acquired with our procedure generalizes across pretrained networks, allowing flexibility in choosing the final model or updating it as newer versions get released.
\end{abstract}

\section{Introduction}
\label{Sec:Introduction}
Text classification has a long history and numerous applications in the field of natural language processing \citep{jindal,lai}. Since the debut of Transformers \citep{transformers}, transfer learning using Large Language Models (LLMs) such as BERT, RoBERTa, and ELECTRA has become increasingly popular among practitioners \citep{devlin,roberta, electra}. fine-tuning these models on text classification datasets including GLUE, MultiNLI, and IMDb significantly improved their state-of-the-art performance \citep{howard,devlin}. However, in many practical scenarios, downstream text datasets are either scarcely labeled or are unlabeled at all, restricting supervised transfer learning. At the same time, manual labeling is often laborious and costly, which calls for a careful and targeted selection of examples for annotation using techniques such as active learning \citep{schroeder}. The standard active learning pipeline iterates over the following steps: (1) train a model on the labeled subset of data, (2) query this model to select unlabeled samples for annotation, and (3) label chosen samples and move them to the labeled split \citep{lewis}. When used together with transfer learning from LLMs, this procedure requires sequentially re-fine-tuning models with up to billions of parameters, which is very expensive, if feasible at all.

\begin{figure*}[!htp]
\centering
\begin{minipage}{0.43\textwidth}
\begin{center}
\includegraphics[scale=0.5]{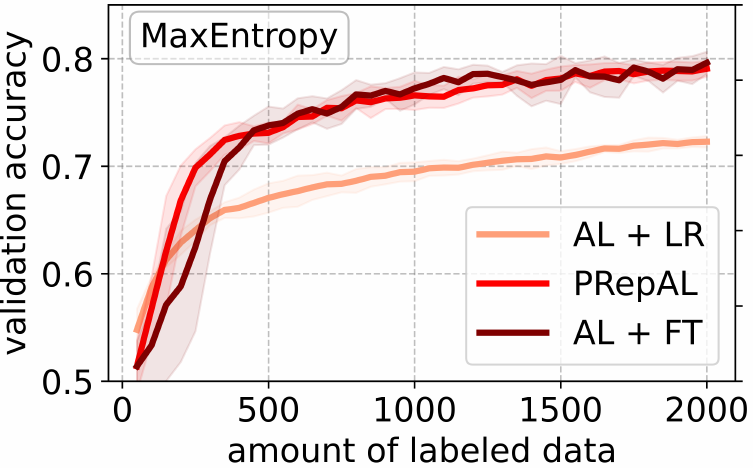}
\end{center}
\end{minipage}
\begin{minipage}{0.56\textwidth}  
\begin{center}
\begin{tabular}{llll}
\toprule
Phase & AL+LR & PRepAL & AL+FT\\
\midrule
Precomputation (s) & 261 & 261 & --- \\
Tot. Retraining (s) & 5 & 5 & 79K \\
Tot. Acquisition (s) & 356 & 356 & 12K \\
Final training (s) & 0.9 & 7K & 7K\\
\midrule
Total (s) & 0.5K & 7.5K & 98K\\
\midrule
Accuracy (\%) & 72.27 & 79.05 & 79.60 \\
\bottomrule
\end{tabular}
\end{center}
\end{minipage}
\caption{Active learning with MaxEntropy acquisition function and BERT backbone on QNLI across different strategies over $39$ labeling iterations. \textbf{Left:} validation performance after training on labeled data thus far. Error bands represent $\pm 1$ standard deviation. \textbf{Right:} wall-clock time (in seconds) spent on each phase and validation accuracy of the final model trained on $2,000$ acquired samples. All models trained to convergence on five cores and a Tesla V100-SXM2-32GB GPU.}
\label{Fig:SNEAK}
\end{figure*}

Recently published studies attempt to mitigate this problem by either using just a single active learning iteration, introducing additional proxy-models, or carefully synchronizing model retraining and human labeling \citep{xie, sanh, famie}. These methods, however, are either impractical, lack in performance, require considerable memory overhead, or suffer from all of the above. In this work, we propose an alternative approach to transfer learning with active learning that brings extraordinary speedups while avoiding these deficiencies. 

\paragraph{Our method.} We introduce \emph{PRepAL (Pretrained Representation Active Learning)}, which precomputes data representations using a pretrained LLM and, on each active learning (AL) iteration, fits a simple linear classifier, e.g., Logistic Regression (LR), avoiding resource-consuming fine-tuning (FT) until the desired amount of data is labeled. This simple procedure yields surprisingly competitive performance with negligible additional resources and time per active learning iteration. For example, fitting Logistic Regression on $2,000$ QNLI representations extracted from a pretrained BERT model takes $0.2$ seconds, whereas fine-tuning the entire BERT model takes nearly two hours. Thus, total time for model retraining in the $39$-step active learning procedure in Figure \ref{Fig:SNEAK} is just over $5$ seconds with PRepAL, which is three orders of magnitude less than with standard fine-tuning (5 seconds vs. 79K seconds or 22 hours). In practice, this speedup helps avoid costly delays between active learning iterations associated with model retraining, allowing human annotators to complete all labeling in one sitting. At the same time, the quality of annotated data remains high: $2,000$ labeled QNLI samples acquired through PRepAL with MaxEntropy scoring function show only $0.55\%$ drop in validation accuracy of the final fine-tuned BERT model compared to those selected with MaxEntropy and fine-tuning. The efficient retraining routine of PRepAL allows for sequential data labeling as opposed to batching and, as our experiments in Section \ref{Sec:Experiments} demonstrate, this can further improve data quality in the early stages of active learning. The data acquired by PRepAL using one LLM as a backbone can successfully fine-tune a different pretrained LLM, as our experiments with BERT and RoBERTa indicate; this transferability allows switching between final model architectures without rerunning active learning. PRepAL can operate with virtually any acquisition function and, hence, is a general mechanism that improves the efficiency of active learning.\\

The remainder of the paper is organized as follows: in Section \ref{Sec:RelatedWork}, we discuss common active learning strategies and other related work. Section \ref{Sec:Method} gives a detailed description of PRepAL. Section \ref{Sec:Experiments} showcases its performance on classic text classification datasets in conjuction with different active learning methods. Sections \ref{Sec:Discussion} and \ref{Sec:Limitations} close the paper with discussion on PRepAL, its strengths and limitations, and offer avenues for future work.

\section{Related Work}
\label{Sec:RelatedWork}
The surge of interest in active learning over the past few decades inspired a wealth of literature surveys \citep{settles,fu,aggarwal,zhan}. Like \citet{schroeder} and \citet{zhan}, we focus on pool-based active learners as they are the most prevalent and are natural for text classification tasks. Active learning algorithms of this type have access to the entire pool of unlabeled data and make decisions via ranking samples with an ad-hoc acquisition function $A(x)$ \citep{xie}. We follow a recent taxonomy by \citet{schroeder} designed specifically for deep learning and describe several relevant active learning methods in more detail.

\paragraph{Data-based.} The methods in this category focus on the unlabeled data itself and are the most model-agnostic. Designed primarily for convolutional neural networks, CoreSet by \citet{coreset} acquires unlabeled data in a greedy manner as to best cover the dataset manifold within the representation space. That is, CoreSet selects instances that maximize the acquisition function $A(x)=\min_{x_j\in L}\lVert \Phi(x)-\Phi(x_j)\rVert_2$ where $L$ is the current labeled dataset and where $\Phi$ is the current embedding mapping. A handful of methods enforce representativeness of selected samples through clustering. \citet{nguyen} use K-medoid clustering for sample density estimation; \citet{xu} run K-means within the SVM margin and send cluster centroids for annotation. \citet{perez} send entire clusters for inspection and labeling by a human.

\paragraph{Model-based.} These methods rely on features of the learner. \citet{settles_egl} was first to use expected norm of the loss gradient with respect to learner's parameters to assess the potential influence of any unlabeled sample on training. This algorithm and its close adaptations are commonly referred to as Expected Gradient Length (EGL). \citet{huang} apply EGL for speech recognition and discuss it from a variance reduction perspective. Formally, the acquisition function of EGL is $A(x)=\mathbb{E}_{\hat{y}\sim \hat{p}(y|x)}\lVert\nabla_{\theta}\ell(x,\hat{y},\hat{\theta})\rVert_2^2$ where $\ell$ is the loss function, $\hat{\theta}$ are the current model parameters, and $\hat{p}(y|x)$ is the estimate of class probabilities at the unlabeled sample $x$. Applying convolutional neural networks for text classification, \citet{zhang} score unlabeled samples by the length of the embedding space update weighted by the current class probability estimates as above. \citet{tong} take a margin-based approach and choose unlabeled points that lie closest to the decision hyperplane of SVM. \citet{ducoffe} extend this idea to deep neural networks by choosing adversarial examples instead. \citet{gissin} develop Discriminative Active Learning (DAL); they fit a separate classifier on the learned representations of the data that discriminates between labeled and unlabeled instances, and acquire those predicted unlabeled with higher confidence. \citet{goral} introduce Goal-Oriented Active Learning (GORAL), which uses influence functions to approximate utility of labeling any datapoint with respect to a particular objective, e.g., negative validation loss or negative prediction entropy.

\paragraph{Prediction-based} The algorithms in this subclass utilize predictions of the current learner (ensemble of learners) to guide acquisition. A large body of studies choose to label samples with the maximum uncertainty as expressed by the model. In the context of text classification, \citet{lewis} measure uncertainty as entropy of the current class probabilities, i.e., $A(x)=\mathbb{H}(\hat{p}(y|x))$. \citet{beluch} find that variation ratio $A(x)=1-\max_i\hat{p}(y_i|x)$, originally introduced by \citet{freeman}, is competitive for active learning in image classification. \citet{houlsby} propose Bayesian Active Learning via Disagreement (BALD) and estimate uncertainty as the mutual information between the predictions and the parameters of a Bayesian model, which they reformulate as $A(x)=\mathbb{H}(y|x)-\mathbb{E}_{ p(\omega)}\mathbb{H}(y|\omega,x)$. \citet{gal} model the posterior $p(\omega)$ as the dropout distribution for BALD when applied in the image classification domain. BALD may overestimate the mutual information between a batch of unlabeled samples and model parameters, making it less effective in batch-mode acquisition. Accounting for this shortcoming, \citet{kirsch} introduce BatchBALD that scores selected points jointly with $A(X)=\mathbb{H}(Y)-\mathbb{E}_{p(\omega)}\mathbb{H}(Y|\omega,X)$ where $X=\{x_1,x_2,\ldots,x_n\}$ and $Y=\{y_1,y_2,\ldots,y_n\}$.

\paragraph{Active learning with proxy models.} Akin to our study, a handful of works are concerned with accelerating the active learning routine common to all of the techniques above. \citet{xie} utilize pretrained feature embeddings for one-shot label querying; however, this method is inferior to many existing baselines that iteratively retrain classifiers on newly labeled data and was evaluated on images only. \citet{shelmanov} retrain a less bulky proxy model---DistillBERT \citep{sanh}---within the active learning loop and fine-tune BERT once on the final labeled dataset. This algorithm exhibits a discrepancy between BERT performance on the approximate and the baseline labeled datasets while bringing only marginal computational savings since DistillBERT has just 40\% less parameters than BERT itself. \citet{coleman} adhere to a similar approach by reducing the width and depth of the full models, which again does not enjoy nearly as much compute efficiency as our method. Like us, \citet{jiao} use Logistic Regression on top of embeddings extracted from a pretrained model during data labeling but apply their method in a very specific medical imaging domain and only with entropy-based acquisition functions. \citet{famie} resort to retraining both the main LLM and the proxy MiniLM \citep{wang} on each active learning iteration but do so in parallel with the current annotation step to save time. This method is even more computationally expensive than the original active learning routine and requires accurate synchronization between labeling and training to realize potential speedups. In contrast, our approach is general, conceptually simple, requires no additional proxy models, computationally cheap on each iteration, and incurs negligible performance drop, if any (Section \ref{Sec:Experiments}).

\section{Method}
\label{Sec:Method}

\begin{table*}[hb]
\caption{Text classification datasets used in this study. The number of classes is denoted by $\text{K}$. AG News was downsampled to $12,500$ documents per class due to limited resources and time.\\}
\centering
\begin{tabular}{lllllll}
\toprule
Dataset   & Task & Size  & $\text{K}$  & Prior & Source & License  \\
\midrule
SST-2 & sentiment analysis & $67,349$ & $2$ & $56\%$ & \citet{sst2} & CC0\\
CoLa & acceptability & $8,551$ & $2$ & $70\%$ & \citet{cola} & CC0 \\
QNLI & question-answering & $102,671$   & $2$  & $50\%$ &  \citet{glue} & CC-BY-SA 4.0\\
IMDb & sentiment analysis & $25,000$ & $2$ & $50\%$ & \citet{imdb} & ---\\
AG News & categorization & $50,000$ & $4$ & uniform & \citet{agnews} & ---\\
\bottomrule
\end{tabular}
\label{Table:Datasets}
\end{table*}

\begin{algorithm}[!t]
\SetAlgoLined
\SetKwInOut{Input}{Input}
\Input{Acquisition function $A$, active learning iterations $T$ and acquisition batch size $b$, classification dataset $\mathcal{D}=\{(X_i,Y_i)\}_{i=1}^n$, initially labeled indices $I_0\subset[n]$, pretrained LLM $\hat{\Phi}$, classifier $\psi$, loss function $\mathcal{L}$.}
\BlankLine
Precompute representations: $\tilde{X}\leftarrow\hat{\Phi}(X)$\;
\For{$t\in [T]$}{
$\psi_t\leftarrow \argmin_{\psi}\mathcal{L}(\psi(\tilde{X}_{I_{t-1}}), Y_{I_{t-1}})$\;
$I\leftarrow\text{Top}_b\{i\notin I_{t-1},\text{key}=A(X_i,\psi_t)\}$\;
$I_t\leftarrow I_{t-1}\cup I$\;
}
Fine-tune LLM: $\mu\leftarrow\argmin_{\psi\circ\Phi}\mathcal{L}(\psi(\Phi(X_{I_T})), Y_{I_T})$\;
\SetKwInOut{Return}{Return} 
\Return{$\mu$}
\caption{Pretrained Representation Active Learning (PRepAL)}
\label{Alg:PRepAL}
\end{algorithm}

As discussed in Section \ref{Sec:RelatedWork}, previous works have used proxy models for data acquisition during active learning, hoping that labeled subsets transfer to more powerful but less efficient training strategies like fine-tuning LLMs. Our method, PRepAL, follows the same paradigm but is simpler and more efficient. Given a pretrained LLM $\hat{\Phi}$ as the backbone, our method precomputes associated data representations $\hat{\Phi}(X)$ and uses them within the notoriously costly active learning loop, refitting a single-layer classifier $\psi$ (e.g., Logistic Regression) in each active learning iteration (Algorithm \ref{Alg:PRepAL}). PRepAL makes AL iterations magnitudes more affordable and avoids unwanted delays caused by lengthy model retraining. For example, total time spent retraining Logistic Regression on precomputed representations of labeled QNLI samples over $39$ active learning iterations is just $5$ seconds, while all $39$ re-fine-tuning cycles take almost $22$ hours (Figure \ref{Fig:SNEAK}). Most importantly, data labeled using PRepAL can be effectively used to fine-tune that same backbone LLM, achieving the best of both worlds: high efficiency and competitive performance.

PRepAL can be applied in conjunction with virtually any acquisition function; however, not all of them fit its simplified retraining procedure equally well. Acquisition methods that operate on the representation space of the trained model, e.g., CoreSet and DAL, exhibit larger performance gaps when trained on data acquired through PRepAL compared to active learning with re-fine-tuning. These algorithms assume that feature extractors change as a result of retraining with more labeled data, which does not happen with the standard PRepAL pipeline where data representations are precomputed once and used for all active learning iterations. In principle, however, we can vary complexity of the model $\psi$ (e.g., by adding hidden layers) that is retrained on static features $\tilde{X}$, trading off the increase in required resources for the dynamic embedding space and higher acquisition quality. Thus, PRepAL is not limited to any particular backbone LLM, classifier types or acquisition functions; rather, it describes a flexible and universal strategy of using pretrained representations for more efficient active learning where fine-tuning pretrained models is state-of-the-art.

\begin{figure*}[hbt!]
\centering
\includegraphics[width=0.96\linewidth]{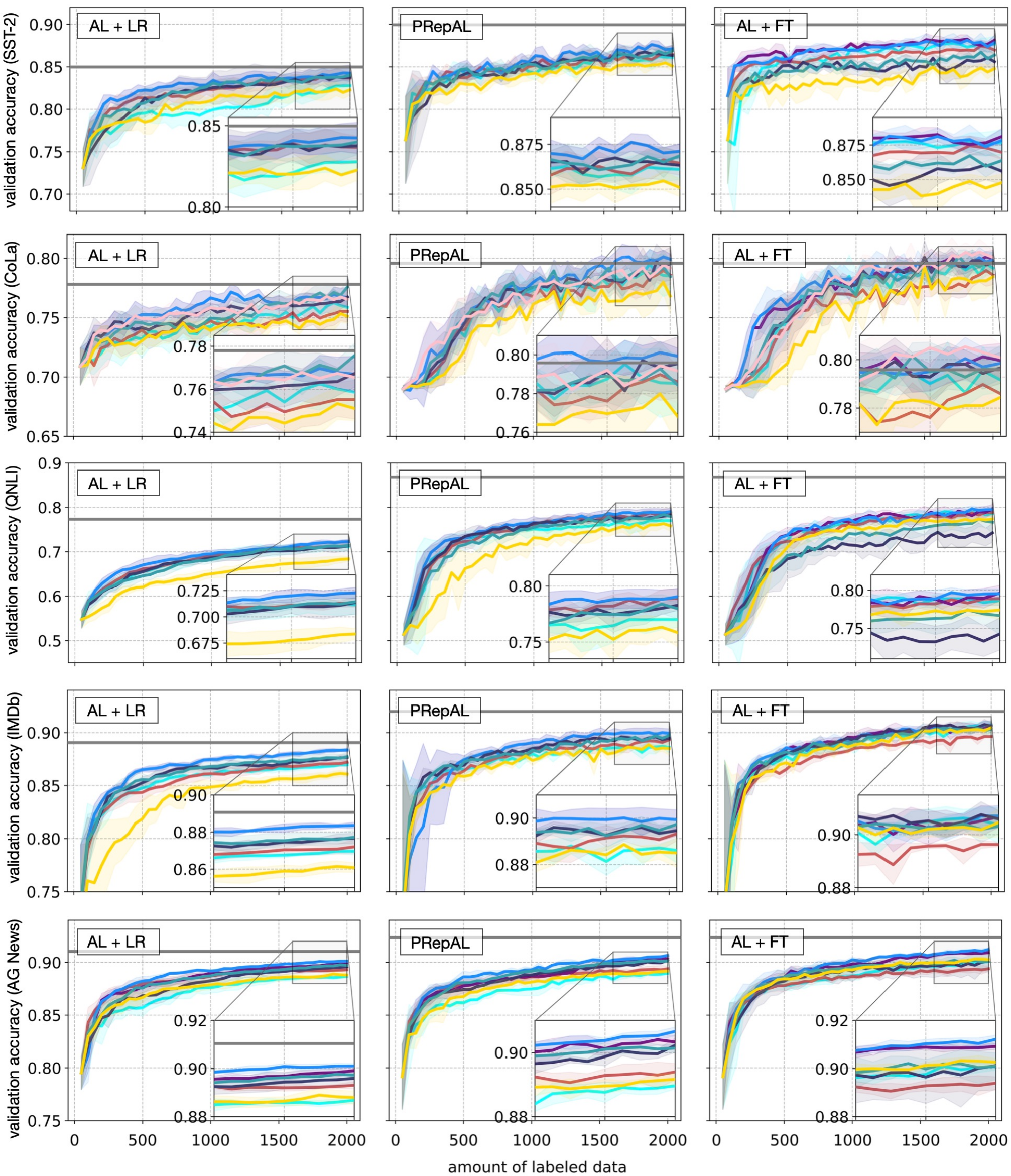}\\\hspace{\fill}\\
\includegraphics[width=0.65\linewidth]{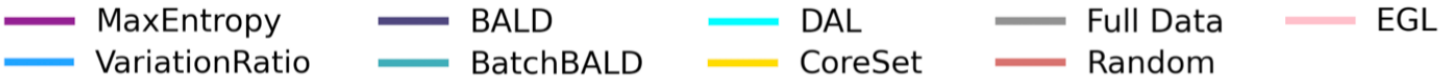}
\caption{Validation accuracy of final models across different acquisition functions, retraining methods, and datasets. All use BERT as the backbone LLM. Error bands represent $\pm 1$ standard deviation.}
\label{Fig:Insets}
\end{figure*}

\begin{figure*}[!hbt]
\centering
\includegraphics[width=0.99\textwidth]{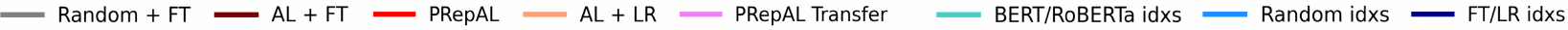}
\begin{subfigure}[b]{\textwidth}
    \centering
    \includegraphics[width=0.9\textwidth]{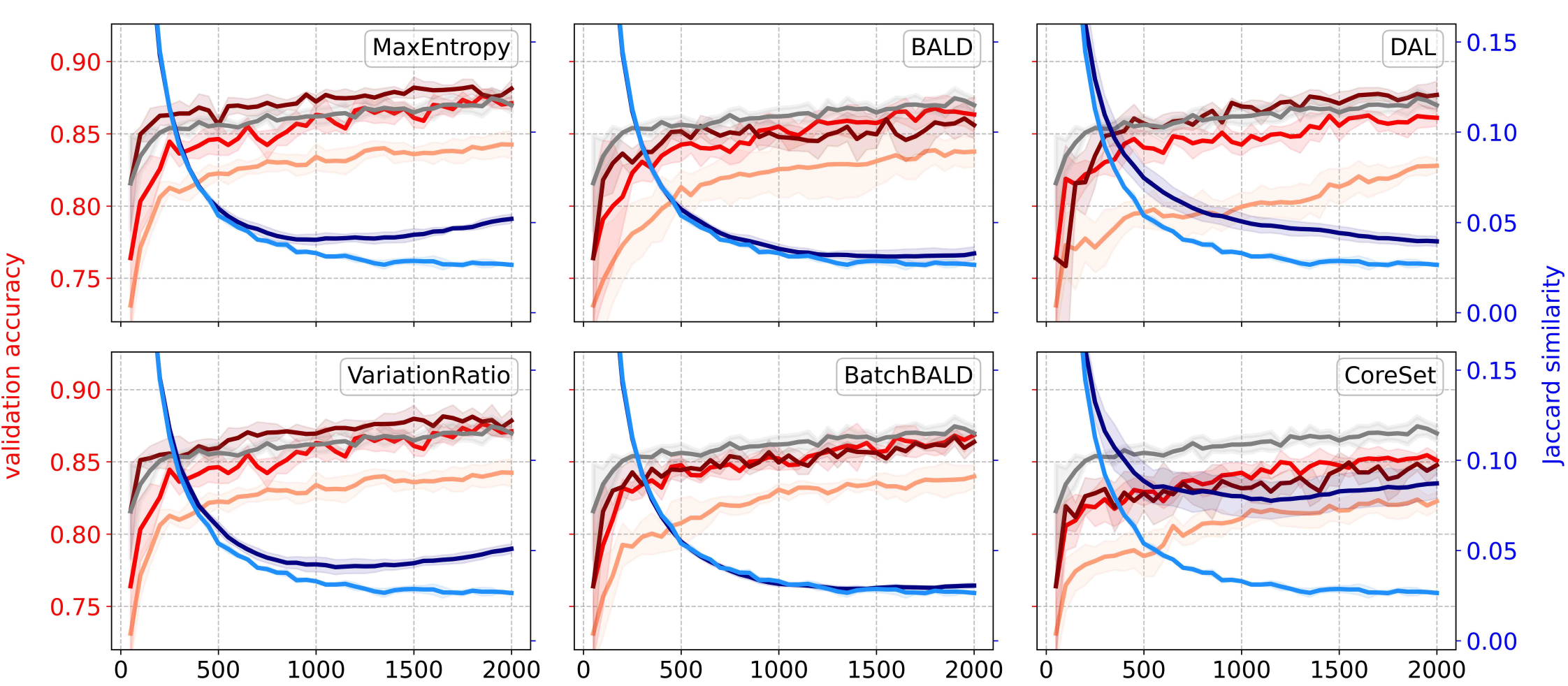}
    \caption{BERT backbone}
    \label{SST-2:bert}
\end{subfigure}
\hfill
\begin{subfigure}[b]{\textwidth}
    \centering
    \includegraphics[width=0.98\textwidth]{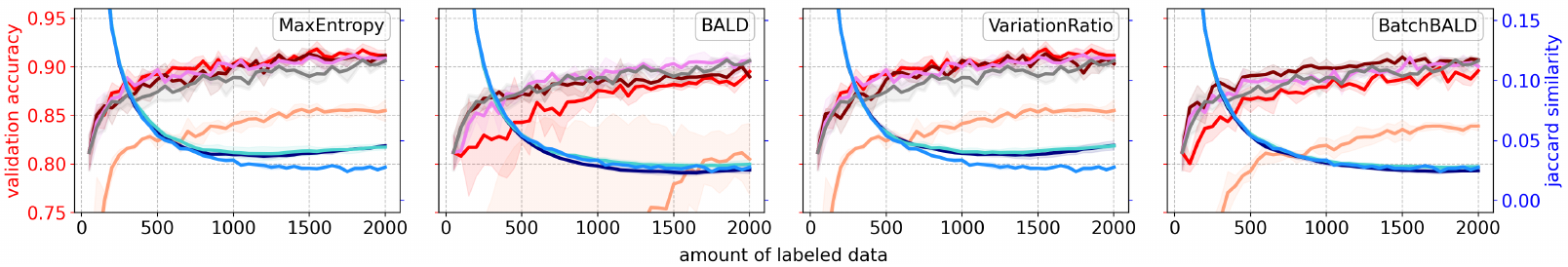}
    \caption{RoBERTa backbone}
    \label{SST-2:roberta}
\end{subfigure}
\caption{SST-2 dataset. The red-toned curves and the grey curve show the validation accuracy of different models with different active learning protocols. The blue-toned curves indicate Jaccard similarity between subsets of data indices selected by different active learning protocols and the data indices selected by AL+FT. Error bands represent $\pm 1$ standard deviation. Among all acquisition functions, only DAL presents a visible performance gap between PRepAL and AL+FT.}
\label{Fig:SST2}
\end{figure*}

\section{Experiments}
\label{Sec:Experiments}
In this section, we present experimental results that benchmark several active learning methods for text classification using pretrained BERT \citep{devlin} and RoBERTa \citep{roberta} to evaluate PRepAL in an extensive ablation study (full results can be found in Appendix \ref{Sec:Appendix}). We perform our evaluation across $8$ different acquisition functions: Random, MaxEntropy, VariationRatio, BALD, BatchBALD, DAL, CoreSet (greedy), and EGL (see Section \ref{Sec:RelatedWork} for details) on $5$ common text classification datasets: SST-2, CoLa, QNLI, IMDb, and AG News (Table \ref{Table:Datasets}). We run EGL only on the smallest dataset (CoLa) due to time constraints. Following \citet{devlin}, we use the Adam optimizer \citep{adam} with cross-entropy criterion for $3$ training epochs, batch size of $16$, dropout rate of $0.1$, early stopping, no weight decay, and 1e-6 and 2e-5 as learning rates of the backbone and the linear classifier, respectively. Each experiment starts with $50$ randomly chosen labeled samples and acquires $50$ more on each of the $39$ subsequent AL iterations. Precomputed data representations are extracted from the last hidden state of the backbone LLM and are reduced using average pooling (we tested other feature extraction setups but preliminary runs showed little dependence on these hyperparameters). All experiments were set up in PyTorch \citep{pytorch} and PyTorch Lightning \citep{falcon}, run on an internal cluster with Tesla V100-SXM2-32GB GPUs installed, and repeated across five random seeds, taking approximately $12,500$ GPU hours total.

\begin{figure*}[!hbt]
\centering
\includegraphics[width=0.99\textwidth]{handles.pdf}
\begin{subfigure}[b]{\textwidth}
    \centering
    \includegraphics[width=0.9\linewidth]{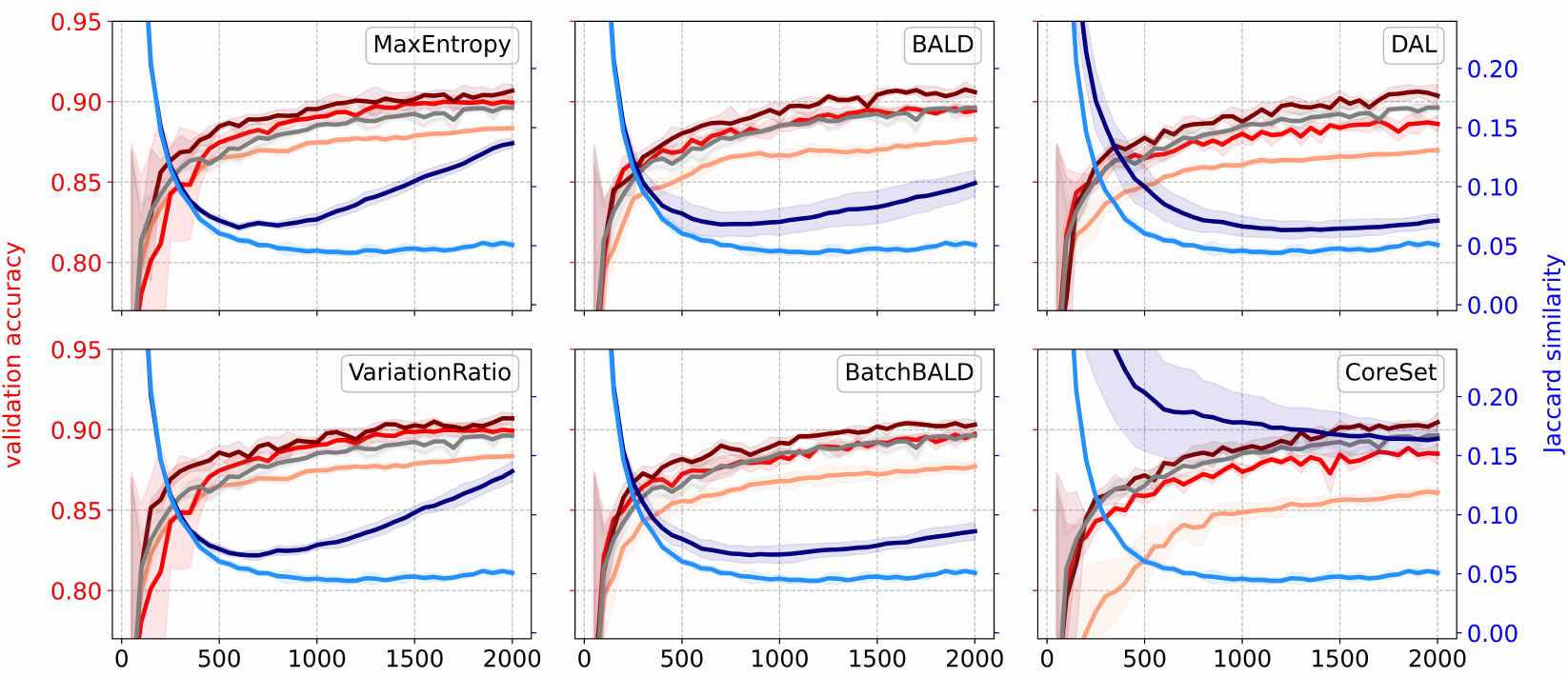}
    \caption{BERT backbone}
\end{subfigure}
\hfill
\begin{subfigure}[b]{\textwidth}
    \centering
    \includegraphics[width=0.98\textwidth]{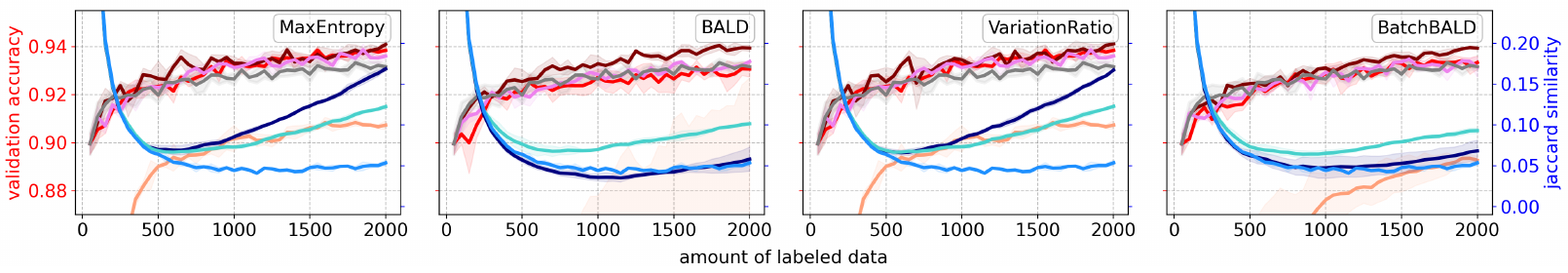}
    \caption{RoBERTa backbone}
    \label{IMDb:roberta}
\end{subfigure}
\caption{IMDb dataset. The red-toned curves and the grey curve show the validation accuracy of different models with different active learning protocols. The blue-toned curves indicate Jaccard similarity between subsets of data indices selected by different active learning protocols and the data indices selected by AL+FT. Error bands represent $\pm 1$ standard deviation. PRepAL retains the above-random performance of AL+FT with MaxEntropy and VariationRatio.}
\label{Fig:IMDb}
\end{figure*}

\paragraph{Active learning protocols.} For each dataset and acquisition function pair, we implement three active learning strategies---AL+LR, PRepAL, and AL+FT---that differ in the model retraining procedure during and after active learning. In particular, AL+LR and PRepAL refit a Logistic Regression model on each data labeling iteration, while AL+FT re-fine-tunes the entire LLM from its original pretrained parameters together with a linear classification head. Thus, discounting for randomness, data selected with AL+LR and PRepAL must be identically the same. When the desired amount of data is acquired, AL+LR still fits a Logistic Regression while PRepAL and AL+FT fine-tune the original pretrained LLM. The validation accuracy of these final models is recorded in Figures \ref{Fig:SST2} and \ref{Fig:IMDb}. Figures \ref{SST-2:roberta} and \ref{IMDb:roberta}, additionally show how well data acquired by refitting a Logistic Regression model on top of BERT representations transfers to fine-tune a RoBERTa model (PRepAL Transfer).

Figure \ref{Fig:Insets} depicts relative performance of different active learners across datasets and retraining procedures. A similar study was carried out by \citet{ibm} and, although their setup is slightly different, their results match ours where comparison is appropriate. In particular, we observe that the simplest uncertainty-based acquisition functions---MaxEntropy and VariationRatio---perform consistently well and are superior to other methods in most scenarios. Among others, only DAL consistently outperforms random acquisition when used together with the most powerful active learning protocol, AL+FT. Note, on the other hand, that DAL and especially CoreSet lose to Random in practically all setups when Logistic Regression is used for querying samples for acquisition. Indeed, as discussed in Section \ref{Sec:Method}, these methods are not expected to shine when data representations remain unchanged across labeling iterations, as is the case with AL+LR and PRepAL, which share the same acquisition strategy. Figure \ref{Fig:Insets} shows that it is crucial to fine-tune the final LLM to achieve best performance, regardless of the quantity of available labeled data and of the way it was obtained. This is more clearly shown in Figures \ref{Fig:SST2} and \ref{Fig:IMDb}.

\begin{table*}[!b]
\caption{Validation accuracy (mean$\pm$std, in $\%$) of the final models fine-tuned on $2,000$ labeled samples. Accuracy above random labeling is shown in bold.}
\centering
\small
\begin{tabular}{ll|lll|lll}
\toprule
\multirow{2}{*}{Algorithm} & \multirow{2}{*}{Protocol} & \multicolumn{3}{|c|}{BERT} & \multicolumn{3}{c}{RoBERTa}\\
& & \multicolumn{1}{c}{SST-2} & \multicolumn{1}{c}{IMDb} & \multicolumn{1}{c}{CoLa} & \multicolumn{1}{c}{SST-2} & \multicolumn{1}{c}{IMDb} & \multicolumn{1}{c}{CoLa}\\
\midrule
Random & AL+FT & $86.9\pm0.5$ & $89.6\pm0.1$ & $78.6\pm0.8$ & $90.6\pm0.3$ & $93.1\pm0.1$ & $80.4\pm0.5$ \\
\midrule
\multirow{4}{*}{MaxEntropy}
& AL+LR & $84.2\pm0.9$ & $88.3\pm0.1$ & $76.7\pm0.8$ & $85.5\pm1.1$ & $90.7\pm0.2$ & $80.7\pm0.2$ \\
& PRepAL & $\mathbf{87.1\pm0.4}$ & $\mathbf{89.9\pm0.4}$ & $\mathbf{79.9\pm0.4}$ & $\mathbf{91.1\pm0.5}$ & $\mathbf{93.8\pm0.2}$ & $\mathbf{81.4\pm1.4}$\\
& AL+FT & $\mathbf{88.1\pm0.6}$ & $\mathbf{90.7\pm0.4}$ & $\mathbf{80.0\pm0.7}$ & $\mathbf{91.1\pm0.1}$ & $\mathbf{94.1\pm0.2}$ & $\mathbf{80.8\pm0.1}$\\
& Transfer & --- & --- & --- & $\mathbf{90.9\pm0.4}$ & $\mathbf{93.6\pm0.1}$ & $\mathbf{81.3\pm0.2}$\\
\midrule
\multirow{4}{*}{BatchBALD}
 & AL+LR & $84.0\pm0.7$ & $87.7\pm0.4$ & $77.6\pm0.7$ & $83.9\pm0.4$ & $89.3\pm0.2$ & $79.4\pm0.5$\\
& PRepAL & $\mathbf{86.9\pm0.2}$ & $\mathbf{89.7\pm0.2}$ & $\mathbf{79.2\pm0.8}$ & $89.6\pm0.5$ & $\mathbf{93.3\pm0.3}$ & $79.2\pm2.5$\\
& AL+FT & $86.3\pm0.5$ & $\mathbf{90.3\pm0.3}$ & $\mathbf{79.3\pm1.1}$ & $\mathbf{90.7\pm0.2}$ & $\mathbf{93.9\pm0.1}$ & $\mathbf{81.4\pm0.3}$\\
& Transfer & --- & --- & --- & $90.1\pm0.6$ & $\mathbf{93.1\pm0.4}$ & $79.7\pm0.7$\\
\bottomrule
\end{tabular}
\label{Table:Result}
\end{table*}

\paragraph{Main results.} In general, PRepAL incurs minimal, if any, accuracy drops compared to AL+FT. When used with the most consistent and successful acquisition functions such as MaxEntropy and VariationRatio, PRepAL tends to close this performance gap as more data becomes available. AL+LR trails the other two strategies by quite a margin, suggesting that investing resources in post-labeling fine-tuning is essential to achieve best performance. We observe that random acquisition is a strong baseline, beating AL+FT with BALD and BatchBALD acquisition functions on several occasions (Figure \ref{SST-2:bert}). Still, PRepAL is almost always better than random labeling (Table \ref{Table:Result}).

\paragraph{Selected data.} Jaccard similarity shows a considerable overlap between data indices selected for labeling through querying Logistic Regression models and those obtained via fine-tuning entire backbone models, which is especially pronounced with MaxEntropy and VariationRatio (the blue-toned curves in Figures \ref{Fig:SST2} and \ref{Fig:IMDb}). Most importantly, in these cases, Jaccard similarity grows steadily with active learning iterations, indicating that PRepAL and AL+FT consistently select similar samples for labeling, which contributes to their even performance. While DAL and CoreSet also exhibit higher than random Jaccard similarity between PRepAL and AL+FT indices, it is only observed over initial iterations and shrinks with more labeled data. This phenomenon may be an artifact of PRepAL's immutable data representations, which these two acquisition methods heavily rely on. We hypothesize that BERT's representation space does not change as much in the beginning of active learning, causing larger overlap between indices selected by PRepAL and by AL+FT in the first few iterations of DAL and CoreSet.

\begin{figure}[h]
\centering
\includegraphics[width=0.38\textwidth]{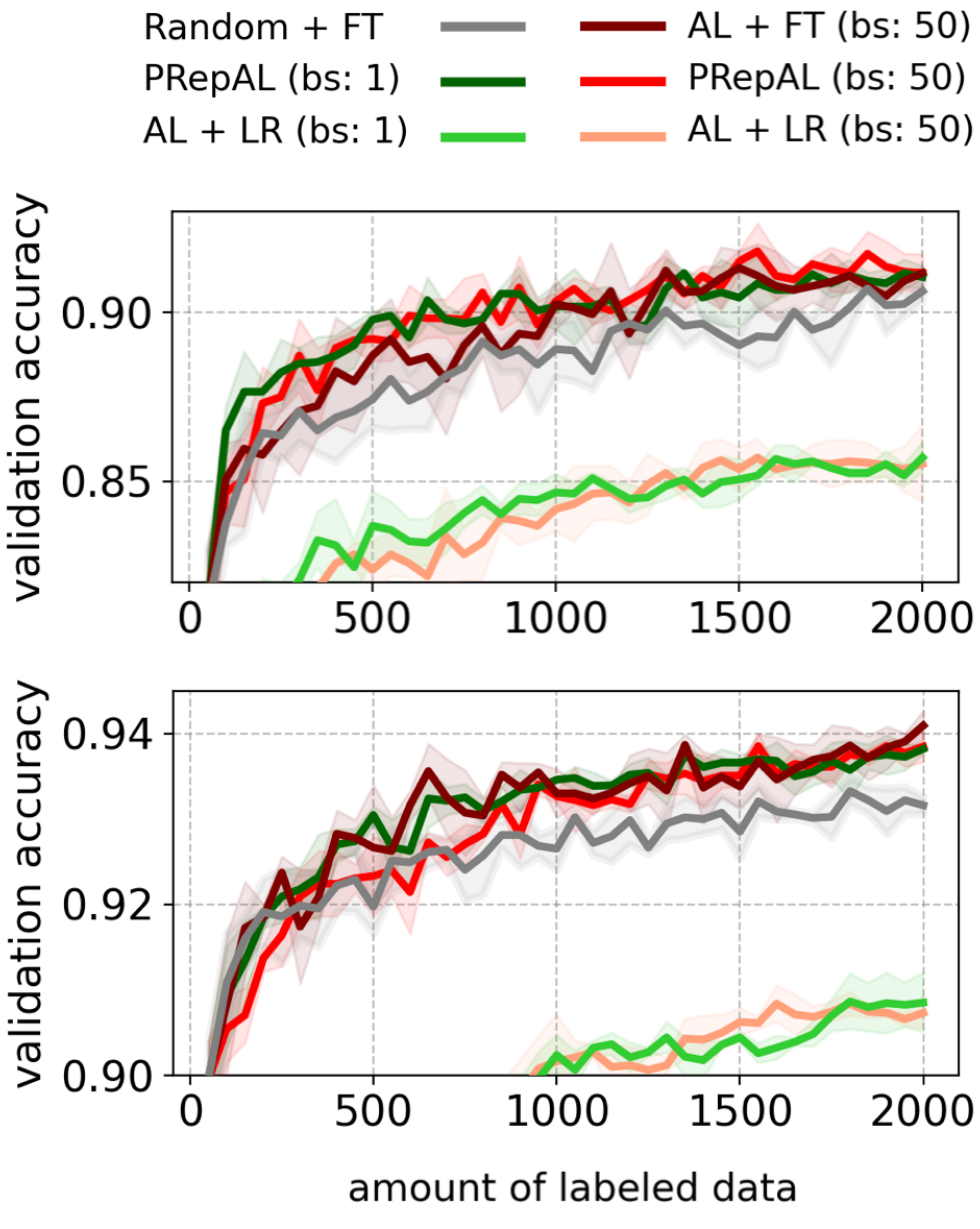}
\caption{Test accuracy of various active learning protocols with MaxEntropy acquisition function and a batch size (bs) of $50$ samples per iteration (red-toned curves), or $1$ sample per iteration (green-toned curves). \textbf{Top:} RoBERTa on SST-2; \textbf{Bottom:} RoBERTa on IMDb.}
\label{Fig:BatchSize}
\vspace{-15px}
\end{figure}

\paragraph{Transferability across models.} Figures \ref{SST-2:roberta} and \ref{IMDb:roberta} show that data acquired by PRepAL with BERT can be successfully transferred to fine-tune a pretrained RoBERTa model (PRepAL Transfer). In fact, across all five datasets and training sizes, it performs no worse than PRepAL with RoBERTa backbone itself, sometimes even surpassing AL+FT (BALD and BatchBALD in Figure \ref{SST-2:roberta}). PRepAL indices transferred from BERT have considerable overlap with those selected by AL+FT using RoBERTa; interestingly, this overlap can be even higher than for PRepAL with RoBERTa itself (Figure \ref{IMDb:roberta}). These observations offer flexibility to choose the final model architecture after PRepAL is initially used. As updated versions of popular large pretrained models are released, one is not required to rerun PRepAL with the new backbone but can instead reuse data labeled previously using a different LLM and achieve a commensurate accuracy.

\paragraph{Reducing the batch size.} One fundamental culprit of most score-based active learning procedures is the lack of diversity in the acquired data due to batching \citep{batch-size-problem, batch-size-short}. Reducing the acquisition batch size often improves model performance \citep{batch-size-1}. However, the extreme costs associated with retraining models  urge practitioners to increase the number of samples acquired on every iteration, sacrificing diversity and, hence, quality of the data. In response, recent works design a variety of algorithms to mitigate the negative consequences associated with large batch acquisition \citep{kirsch, diversity, diversity-2}. The resource efficiency of retraining with PRepAL, on the other hand, allows us to abandon batching whatsoever and acquire training samples one by one---an unthinkable luxury for any other active learning procedure. In Figure \ref{Fig:BatchSize}, we compare the performance of RoBERTa fine-tuned on data acquired using batch sizes of $50$ and $1$ sample(s) per each of $39$ and $1950$ iterations, respectively. Interestingly, we observe that using sequential labeling improves the ultimate model only in the beginning and when labeled data is still limited. This may indicate that, while the diversity within each individual batch is poor, different batches at different iterations are diverse enough to match the quality of the data acquired one by one.

\section{Discussion}
\label{Sec:Discussion}

Motivated to reduce the computational costs of active learning when fine-tuning massive models such as LLMs is state-of-the-art, we propose PRepAL---a universal active learning protocol for quick and memory-efficient acquisition of high-quality data by leveraging pretrained representation spaces. Our method precomputes fixed representations of the unlabeled data using a pretrained LLM and retrains a linear classifier on each active learning iteration in seconds, avoiding unwanted delays between labeling phases. Finally, PRepAL fine-tunes the original LLM on the ultimate subset of labeled data to reach best performance. 

We empirically confirmed the effectiveness of our method using pretrained BERT and RoBERTa models across a variety of text classification datasets and active learning functions. As a byproduct, we benchmarked seven pool-based acquisition methods and found simple uncertainty-based scoring functions like MaxEntropy and VariationRatio to be particularly successful and consistent in this domain. Conveniently, PRepAL is most effective when used together with these functions, showing little performance drop compared to a more laborious data acquisition procedure (AL+FT), in which the entire LLM is re-fine-tuned on each active learning iteration. Fitting just a linear classifier on every active learning iteration allows labeling data points sequentially and not in batches, offering an improvement in quality of the data during the early stages of the active learning loop. The data acquired by the Logistic Regression model not only transfers to fine-tune the original pre-trained backbone architecture but also to other, potentially more advanced models as demonstrated by our experiments with BERT and RoBERTa.

\section{Limitations \& Future Research}
\label{Sec:Limitations} We close the paper by discussing the limitations of our method and sketching the directions for future research. As mentioned in Section \ref{Sec:Method}, not all acquisition functions work equally well with PRepAL. Some methods, like DAL and CoreSet, are more sensitive to having accurate representation spaces, which remain fixed throughout our active learning protocol. On the other hand, in Section \ref{Sec:Experiments} we found these algorithms inferior to other simpler baselines (e.g., MaxEntropy), for which PRepAL matches with its more sophisticated rival AL+FT in terms of the final model performance. In addition, it is trivial to modify our procedure to obtain dynamic representation spaces of retrained models by stacking hidden layers in the classifier attached to BERT and using them for feature extraction. It might be interesting to test whether this adjustment will lead to better performance for acquisition methods like DAL and CoreSet. Further research may explore how viable PRepAL is for other types of downstream tasks such as machine translation or even for applications in computer vision, where fine-tuning deep convolutional networks or Visual Transformers has become a popular practice \citep{imagenet2, vit, imagenet1}. Finally, the flying speed of retraining with PRepAL opens an opportunity to compare batch mode active learning with sequential labeling, which can potentially reveal how exactly the acquisition size impacts data diversity, quality, and the ultimate model performance.

\section{Acknowledgements} The first author was partly supported by the NSF Award 1922658. 

\bibliography{anthology,custom}
\bibliographystyle{acl_natbib}
\appendix
\section{Additional experiments}
\label{Sec:Appendix}
\newpage

\begin{figure*}[!hbt]
\centering
\includegraphics[width=0.99\textwidth]{handles.pdf}
\begin{subfigure}[b]{\textwidth}
    \centering
    \includegraphics[width=0.86\linewidth]{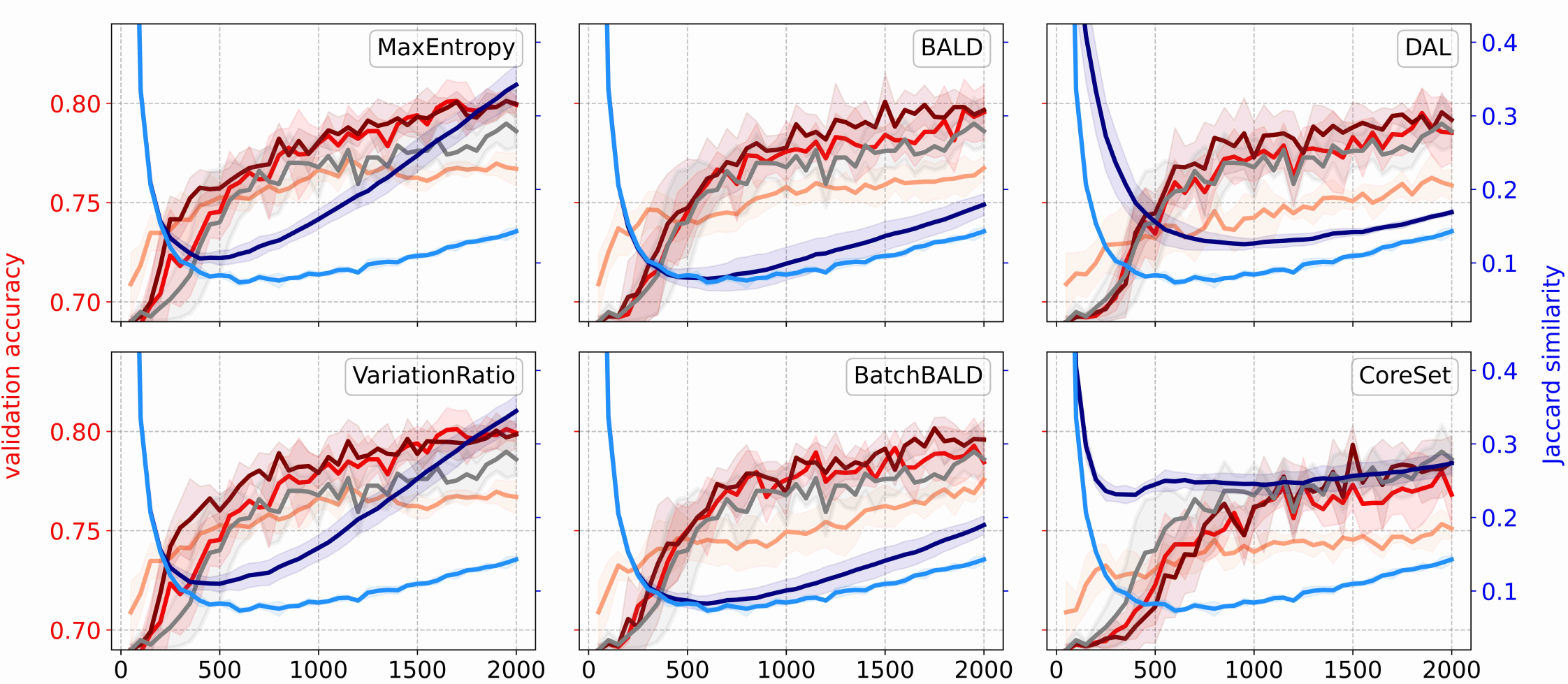}
    \caption{BERT backbone}
\end{subfigure}
\hfill
\begin{subfigure}[b]{\textwidth}
    \centering
    \includegraphics[width=0.98\textwidth]{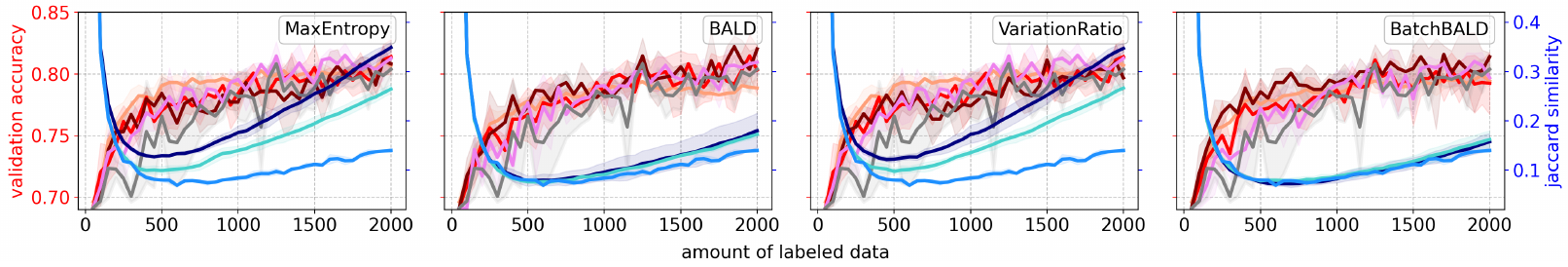}
    \caption{RoBERTa backbone}
    \label{CoLa:roberta}
\end{subfigure}
\caption{CoLa dataset. The red-toned curves and the grey curve show the validation accuracy of different models with different active learning protocols. The blue-toned curves indicate Jaccard similarity between subsets of data indices selected by different active learning protocols and the data indices selected by AL+FT. Error bands represent $\pm 1$ standard deviation. The high values of Jaccard similarity are partly due to the dataset size (only 8, 551 samples). Almost 40\% of the final 2,000 samples selected by AL+FT are in the 2,000 chosen by PRepAL (AL+LR) under MaxEntropy acquisition.}
\label{Fig:CoLa}
\end{figure*}

\begin{figure*}[!hbt]
\centering
\begin{subfigure}[b]{\textwidth}
    \centering
    \includegraphics[width=0.86\linewidth]{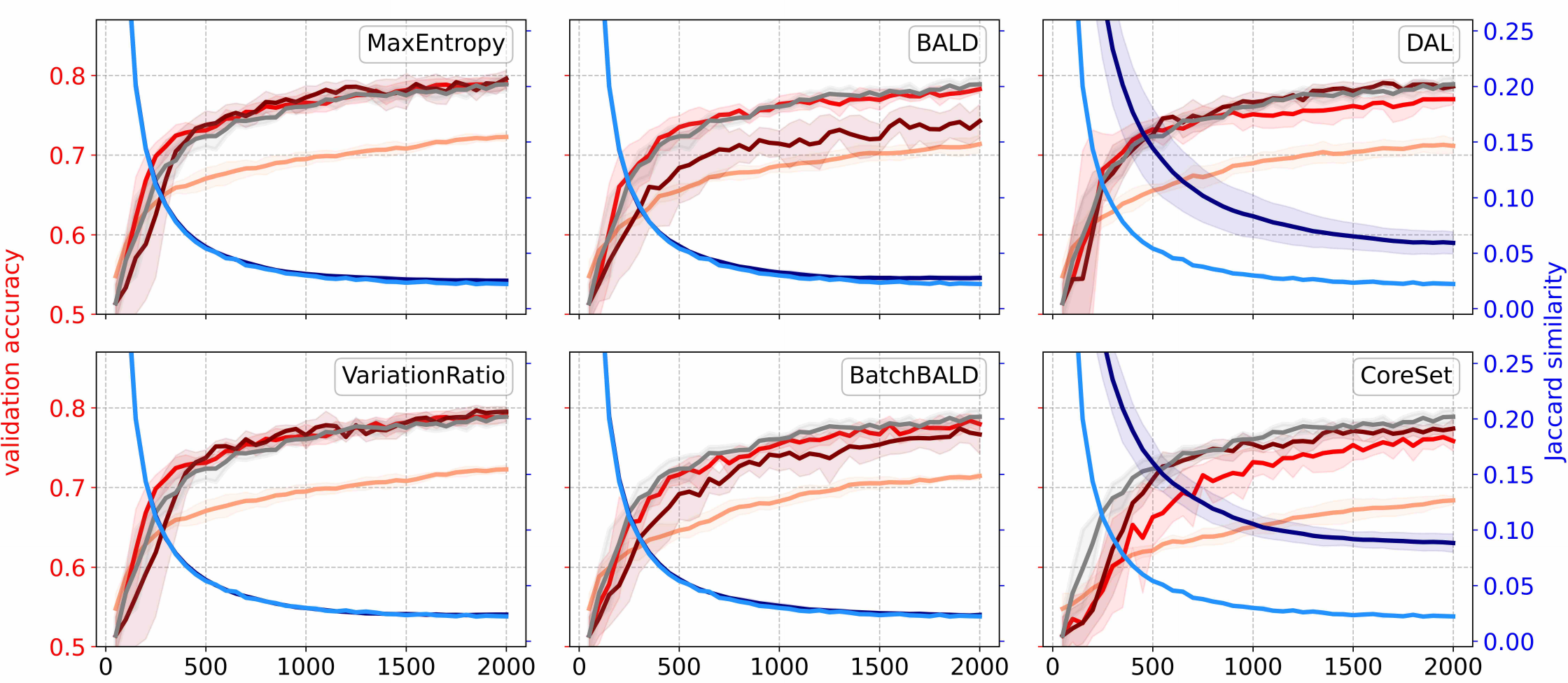}
    \caption{BERT backbone}
\end{subfigure}
\hfill
\begin{subfigure}[b]{\textwidth}
    \centering
    \includegraphics[width=0.98\textwidth]{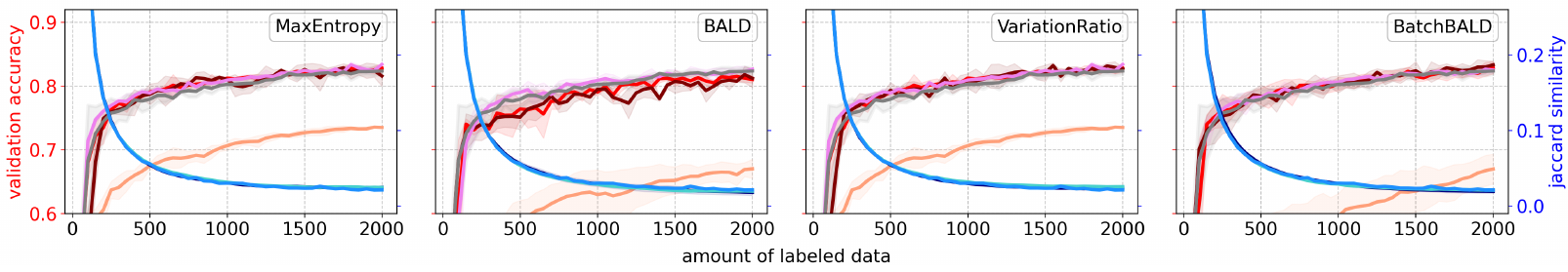}
    \caption{RoBERTa backbone}
    \label{QNLI:roberta}
\end{subfigure}
\caption{QNLI dataset. The red-toned curves and the grey curve show the validation accuracy of different models with different active learning protocols. The blue-toned curves indicate Jaccard similarity between subsets of data indices selected by different active learning protocols and the data indices selected by AL+FT. Error bands represent $\pm 1$ standard deviation. PRepAL and AL+FT perform similarly, with PRepAL at an unexpected advantage for BALD and BatchBALD. Both significantly outperform AL+LR but are not better than random acquisition. The Jaccard similarity between indices associated with PRepAL and AL+FT is indistinguishable from random for uncertainty-based active learners, which is likely due to the dataset size ($100$K$+$ training samples). }
\label{Fig:QNLI}
\end{figure*}

\begin{figure*}[!hbt]
\centering
\includegraphics[width=0.99\textwidth]{handles.pdf}
\begin{subfigure}[b]{\textwidth}
    \centering
    \includegraphics[width=0.86\linewidth]{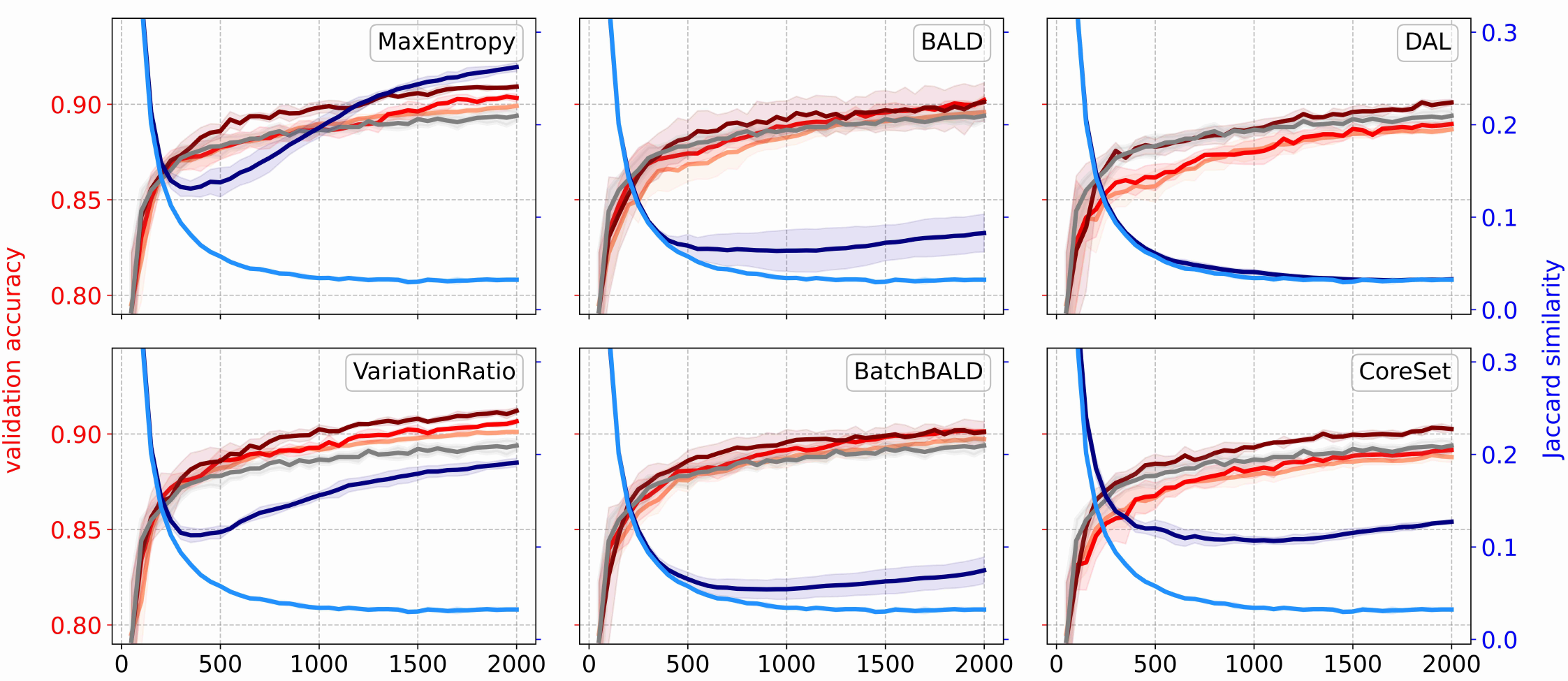}
    \caption{BERT backbone}
\end{subfigure}
\hfill
\begin{subfigure}[b]{\textwidth}
    \centering
    \includegraphics[width=0.98\textwidth]{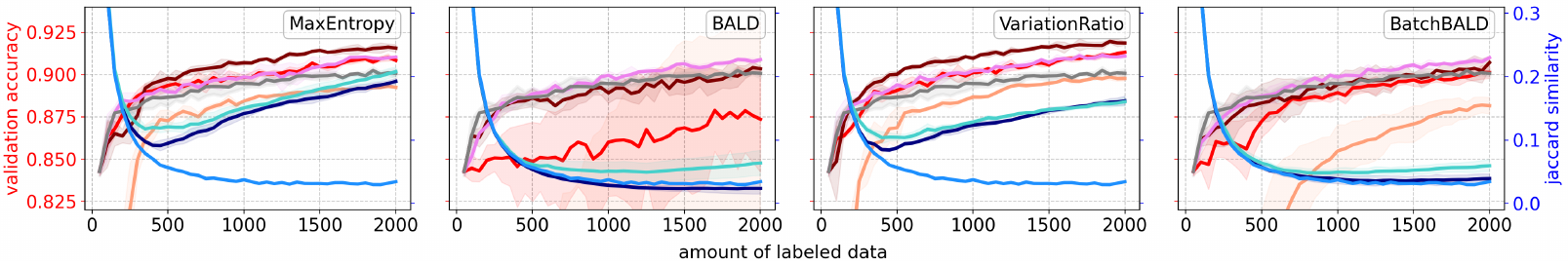}
    \caption{RoBERTa backbone}
    \label{AGNews:roberta}
\end{subfigure}
\caption{AG News dataset. The red-toned curves and the grey curve show the validation accuracy of different models with different active learning protocols. The blue-toned curves indicate Jaccard similarity between subsets of data indices selected by different active learning protocols and the data indices selected by AL+FT. Error bands represent $\pm 1$ standard deviation. Unlike other datasets, AL+LR closes in on the tight performance of PRepAL and AL+FT with BERT and even outperforms random acquisition with subsequent fine-tuning for 4/6 active learning functions.}
\label{Fig:AG News}
\end{figure*}

\begin{table*}[h]
\caption{Validation accuracy (mean$\pm$std, in $\%$) of the final BERT model fine-tuned on $2,000$ labeled samples selected by different acquisition functions with different retraining protocols. Accuracy above random labeling is shown in bold.}
\centering
\small
\begin{tabular}{ll|ccccc}
\toprule
Algorithm & Protocol & SST-2 & QNLI & CoLa & AG News & IMDb\\
\midrule
Random 
& AL+FT & $86.9\pm0.5$ & $78.8\pm0.9$ & $78.6\pm0.8$ & $89.4\pm0.3$ & $89.6\pm0.1$\\
\midrule
\multirow{3}{*}{Max Entropy}
& AL+LR & $84.2\pm0.9$ & $72.2\pm0.5$ & $76.7\pm0.8$ & $\mathbf{89.9\pm0.1}$ & $88.3\pm0.1$\\
& PRepAL & $\mathbf{87.1\pm0.4}$ & $\mathbf{79.0\pm0.7}$ & $\mathbf{79.9\pm0.4}$ & $\mathbf{90.3\pm0.2}$ & $\mathbf{89.9\pm0.4}$\\
& AL+FT & $\mathbf{88.1\pm0.6}$ & $\mathbf{79.6\pm1.0}$ & $\mathbf{80.0\pm0.7}$ & $\mathbf{90.9\pm0.1}$ & $\mathbf{90.7\pm0.4}$\\
\midrule
\multirow{3}{*}{Variation Ratio}
& AL+LR & $84.2\pm0.9$ & $72.2\pm0.5$ & $76.7\pm0.8$ & $\mathbf{90.1\pm0.1}$ & $88.3\pm0.2$\\
& PRepAL & $\mathbf{87.1\pm0.4}$ & $\mathbf{79.0\pm0.7}$ & $\mathbf{79.9\pm0.4}$ & $\mathbf{90.6\pm0.1}$ & $\mathbf{89.9\pm0.4}$\\
& AL+FT & $\mathbf{87.8\pm0.8}$ & $\mathbf{79.5\pm0.6}$ & $\mathbf{79.8\pm0.6}$ & $\mathbf{91.2\pm0.1}$ & $\mathbf{90.7\pm0.3}$\\
\midrule
\multirow{3}{*}{BALD}
& AL+LR & $83.7\pm1.0$ & $71.3\pm1.1$ & $76.7\pm0.6$ & $\mathbf{89.6\pm0.2}$ & $87.6\pm0.3$\\
& PRepAL & $86.3\pm1.2$ & $78.2\pm0.4$ & $\mathbf{79.5\pm1.3}$ & $\mathbf{90.2\pm0.2}$ & $89.5\pm0.2$\\
& AL+FT & $85.6\pm0.9$ & $74.2\pm2.0$ & $\mathbf{79.6\pm0.6}$ & $\mathbf{90.1\pm1.0}$ & $\mathbf{90.6\pm0.3}$\\
\midrule
\multirow{3}{*}{BatchBALD}
& AL+LR & $84.0\pm0.7$ & $71.4\pm0.4$ & $77.6\pm0.6$ & $\mathbf{89.7\pm0.1}$ & $87.7\pm0.4$\\
& PRepAL & $\mathbf{86.9\pm0.2}$ & $77.9\pm0.9$ & $78.4\pm1.3$ & $\mathbf{90.1\pm0.1}$ & $\mathbf{89.7\pm0.2}$\\
& AL+FT & $86.3\pm0.5$ & $76.6\pm2.4$ & $\mathbf{79.6\pm1.0}$ & $\mathbf{90.1\pm0.5}$ & $\mathbf{90.3\pm0.3}$\\
\midrule
\multirow{3}{*}{DAL}
& AL+LR & $82.7\pm0.4$ & $71.1\pm0.9$ & $75.8\pm0.8$ & $88.7\pm0.1$ & $86.9\pm0.2$\\
& PRepAL & $86.1\pm0.5$ & $77.0\pm0.1$ & $78.5\pm1.5$ & $88.9\pm0.1$ & $88.6\pm0.5$\\
& AL+FT & $\mathbf{87.7\pm0.9}$ & $78.6\pm0.5$ & $\mathbf{79.1\pm0.5}$ & $\mathbf{90.1\pm0.1}$ & $\mathbf{90.3\pm0.7}$\\
\midrule
\multirow{3}{*}{CoreSet}
& AL+LR & $82.2\pm0.6$ & $68.3\pm0.5$ & $75.1\pm0.5$ & $88.8\pm0.2$ & $86.1\pm0.2$\\
& PRepAL & $85.1\pm0.5$ & $75.8\pm1.2$ & $76.8\pm1.4$ & $89.2\pm0.2$ & $88.5\pm0.6$\\
& AL+FT & $84.7\pm0.6$ & $77.4\pm1.2$ & $78.4\pm1.3$ & $\mathbf{90.2\pm0.1}$ & $\mathbf{90.4\pm0.5}$\\
\bottomrule
\end{tabular}
\end{table*}

\begin{table*}[h]
\caption{Validation accuracy (mean$\pm$std, in $\%$) of the final RoBERTa model fine-tuned on $2,000$ labeled samples selected by different acquisition functions with different retraining protocols. Accuracy above random labeling is shown in bold.}
\centering
\small
\begin{tabular}{ll|ccccc}
\toprule
Algorithm & Protocol & SST-2 & QNLI & CoLa & AG News & IMDb\\
\midrule
Random & AL+FT & $90.6\pm0.3$ & $82.3\pm0.5$ & $80.4\pm0.5$ & $90.0\pm0.2$ & $93.1\pm0.1$\\
\midrule
\multirow{4}{*}{Max Entropy}
& AL+LR & $85.5\pm1.1$ & $73.5\pm0.2$ & $\mathbf{80.7\pm0.2}$ & $89.2\pm0.1$ & $90.7\pm0.2$\\
& PRepAL & $\mathbf{91.1\pm0.5}$ & $\mathbf{82.8\pm0.6}$ & $\mathbf{81.4\pm1.4}$ & $\mathbf{90.8\pm0.0}$ & $\mathbf{93.8\pm0.2}$\\
& AL+FT & $\mathbf{91.1\pm0.1}$ & $81.6\pm1.6$ & $\mathbf{80.8\pm0.1}$ & $\mathbf{91.5\pm0.2}$ & $\mathbf{94.1\pm0.2}$\\
& Transfer & $\mathbf{90.9\pm0.4}$ & $\mathbf{83.4\pm0.3}$ & $\mathbf{81.3\pm0.2}$ & $\mathbf{91.0\pm0.1}$ & $\mathbf{93.6\pm0.1}$\\
\midrule
\multirow{4}{*}{Variation Ratio}
& AL+LR & $85.5\pm1.1$ & $73.5\pm0.2$ & $\mathbf{80.7\pm0.2}$ & $89.8\pm0.1$ & $90.7\pm0.2$\\
& PRepAL & $\mathbf{91.1\pm0.5}$ & $\mathbf{82.8\pm0.6}$ & $\mathbf{81.4\pm1.4}$ & $\mathbf{91.3\pm0.1}$ & $\mathbf{93.8\pm0.2}$\\
& AL+FT & $90.3\pm0.6$ & $\mathbf{82.6\pm0.2}$ & $79.7\pm1.1$ & $\mathbf{91.8\pm0.1}$ & $\mathbf{94.1\pm0.2}$\\
& Transfer & $\mathbf{90.9\pm0.3}$ & $\mathbf{83.4\pm0.3}$ & $\mathbf{81.3\pm0.2}$ & $\mathbf{91.1\pm0.1}$ & $\mathbf{93.6\pm0.1}$\\
\midrule
\multirow{4}{*}{BALD}
& AL+LR & $80.5\pm3.6$ & $67.0\pm1.5$ & $78.9\pm0.7$ & $72.8\pm9.3$ & $81.9\pm9.2$\\
& PRepAL & $89.5\pm0.6$ & $81.1\pm0.5$ & $80.3\pm2.0$ & $87.3\pm3.1$ & $\mathbf{93.1\pm0.3}$\\
& AL+FT & $89.0\pm0.3$ & $81.2\pm1.4$ & $\mathbf{82.0\pm0.1}$ & $\mathbf{90.3\pm0.1}$ & $\mathbf{93.9\pm0.1}$\\
& Transfer & $\mathbf{90.6\pm0.9}$ & $\mathbf{82.6\pm0.1}$ & $\mathbf{81.0\pm1.0}$ & $\mathbf{90.9\pm0.1}$ & $\mathbf{93.4\pm0.1}$\\
\midrule
\multirow{4}{*}{BatchBALD}
& AL+LR & $83.9\pm0.4$ & $67.0\pm2.0$ & $79.4\pm0.5$ & $88.1\pm0.5$ & $89.3\pm0.2$\\
& PRepAL & $89.6\pm0.5$ & $\mathbf{82.9\pm0.9}$ & $79.2\pm2.5$ & $\mathbf{90.1\pm0.3}$ & $\mathbf{93.3\pm0.3}$\\
& AL+FT & $\mathbf{90.7\pm0.2}$ & $\mathbf{83.3\pm0.5}$ & $\mathbf{81.4\pm0.3}$ & $\mathbf{90.7\pm0.4}$ & $\mathbf{93.9\pm0.1}$\\
& Transfer & $90.1\pm0.6$ & $\mathbf{82.5\pm0.4}$ & $79.7\pm0.7$ & $\mathbf{91.0\pm0.1}$ & $\mathbf{93.1\pm0.4}$\\
\bottomrule
\end{tabular}
\end{table*}

\end{document}